%% file: main.tex
\pdfoutput=1

\documentclass[11pt]{article}

\usepackage{acl}

\usepackage{times}
\usepackage{latexsym}
\usepackage{soul}
\usepackage[T1]{fontenc}

\usepackage[utf8]{inputenc}

\usepackage{microtype}

\usepackage{amsmath}
\usepackage{amssymb}
\usepackage{graphicx}

\usepackage{booktabs}
\usepackage{multirow, multicol}
\usepackage{microtype}

\usepackage{amsfonts}
\usepackage{tabularx}
\usepackage{subfigure}
\usepackage{layouts}

\usepackage[ruled,vlined]{algorithm2e}

\newcommand{\taskfullname}{Multimodal Review Helpfulness Prediction}
\newcommand{\taskabbrname}{MRHP}
\newcommand{\modelname}{SANCL}
\newcommand{\csfigheight}{1.1cm}

\newcommand\blfootnote[1]{%
  \begingroup
  \renewcommand\thefootnote{}\footnote{#1}%
  \addtocounter{footnote}{-1}%
  \endgroup
}

\title{SANCL: Multimodal Review Helpfulness Prediction with Selective Attention and Natural Contrastive Learning}

\author{Wei Han$^{\clubsuit\ast}$, Hui Chen$^{\clubsuit}$, Zhen Hai$^{\dagger\diamond}$, Soujanya Poria$^\clubsuit$, Lidong Bing$^{\dagger}$ \\
  $^\clubsuit$ Singapore University of Technology and Design, Singapore\\
  $^\dagger$ DAMO Academy, Alibaba Group \\ 
  \texttt{\{wei\_han, hui\_chen\}@mymail.sutd.edu.sg}\\
  \texttt{sporia@sutd.edu.sg}\\
  \texttt{\{zhen.hai, l.bing\}@alibaba-inc.com}
}

\begin{document}
\maketitle
\input{Contents/Abstract}
\input{Contents/Introduction}
\input{Contents/Related_Work}
\input{Contents/Method}
\input{Contents/Experiments}
\input{Contents/Analysis}

\input{Contents/Conclusion}

\bibliography{anthology,custom}
\bibliographystyle{acl_natbib}

\input{Contents/Appendix}


\end{document}

%% file: Contents/Abstract.tex
\begin{abstract}

With the boom of e-commerce, Multimodal Review Helpfulness Prediction (MRHP), which aims to sort product reviews according to the predicted helpfulness scores has become a research hotspot.
Previous work on this task focuses on attention-based modality fusion, information integration, and relation modeling, which primarily exposes the following drawbacks:
1) the model may fail to capture the really essential information due to its indiscriminate attention formulation; 
2) lack appropriate modeling methods that take full advantage of correlation among provided data. 
In this paper, we propose SANCL: \textbf{S}elective \textbf{A}ttention and \textbf{N}atural \textbf{C}ontrastive \textbf{L}earning for MRHP.
SANCL adopts a probe-based strategy to enforce high attention weights on the regions of greater significance. 
It also constructs a contrastive learning framework based on natural matching properties in the dataset. 
Experimental results on two benchmark datasets with three categories show that SANCL achieves state-of-the-art baseline performance with lower memory consumption.
Our implementation code for this paper can be found at \url{https://github.com/declare-lab/SANCL}.
\end{abstract}

%% file: Contents/Introduction.tex
\section{Introduction}
\blfootnote{$^\ast$ This work was done when Wei was an intern at Alibaba DAMO Academy.}
\blfootnote{$\diamond$ Corresponding author.}

It is unbelievable to witness an e-commerce boom that has transpired over the past decades \citep{vulkan2020economics}. 
In the virtual bazaar, countless deals are made between mutually invisible sellers and customers from time to time, under the administrator's supervision.
For customers, it may be their biggest headache to determine whether they should pay for a good when being overwhelmed by tempting advertisements, as they know quite a little information about a product in face of the seller's meticulous promotions without any external references.
In this situation, reviews in e-shops that can provide justification information, are thus of great value to customers.
However, the quality of reviews under a certain product page can be disparate---many customers are willing to leave informative feedback on the product, while many others arbitrarily write a few words and even paste irrelevant messages in their comments. 
Therefore, from the perspective of online shopping platforms, they would be welcome and attractive to customers if they provide a service that can intelligently filter and place the most helpful reviews at the top position.
The task in the machine learning field to solve this problem is Review Helpfulness Prediction (RHP)~\citep{tang2013context}.

With the thriving of multimodal learning research and the handy accessibility of multimodal data in this Internet era, the latest progress incorporated image (vision modality) information into the review helpfulness prediction (RHP) \citep{liu2021multi} as Multimodal RHP (MRHP).
Although previous work attained excellent results in MRHP, there are still some drawbacks. 
First, the attention mechanism in these works for representation learning follows the most basic setting---it directly computes out the attention scores based on the representation vectors of tokens or sentences, without any further intervention on the obtained weights \cite{fan2019product}.
Generally, the amount of task-related information in each sentence in a given piece of review may vary greatly---since customers usually casually write these reviews and may insert some meaningless words, such as emotional appreciation or complaints that can not benefit the viewers. 
We observed that due to dataset characteristics in the MRHP task, there are cues to help locate those key sentences in the review text. 
Therefore, we proposed a probe-based selective attention mechanism to employ them for better attention results.

Secondly, it has been revealed that the correlation, e.g., the similarity of feature vectors, among multimodal and multi-field data is an essential factor for task modeling~\cite{xu2020reasoning, chen2019multi}.
Nevertheless, existing studies~\cite{xu2020reasoning, liu2021multi} simply quantified them by similarity metrics, such as cosine value or negative Euclidean distance, and conduct the main tasks based on these similarity scores as features.
Though gained appreciative results, we believe they can be better utilized through the contrastive learning scheme to refine the learned representations, which enables the output layer to make more accurate predictions.

In this paper, we propose a novel framework, \modelname, which incorporates these two basic ideas.
We first generate a special ``probe'' mask that highlights the key sentences from the product and review text.  
Thereafter, these masks are inserted into the computation attention modules to help focus more on task-related sentences. 
Then we construct a contrastive learning framework, which harnesses the internal correlations within multimodal data and the fundamental contrastive predictive coding (CPC) model~\citep{oord2018representation}, to learn better multimodal representations for the main task.
The framework is composed of two feature spaces, dubbed \textit{domains}.
Each domain takes a specific combination of projected representations as input, according to their relation types through semantic analysis.
By minimizing the auxiliary contrastive loss, the multimodal and multi-domain representations can be refined with the inherent relations.
Our contribution can be summarized as follows:
\begin{itemize}
    \item We design a selective attention approach, including probe mask generation and mask-based attention computation, for the information aggregation in \taskabbrname~tasks.
    \item We analyze the characteristics and relations in multimodal reviews and formulate a contrastive learning framework to refine the learned representations.
    \item Extensive experiments on three publicly available datasets show our approach achieves state-of-the-art performance with lower memory consumption.
\end{itemize}

%% file: Contents/Related_Work.tex
\section{Related Work}
In this section, we briefly recap some relevant work in the field of review helpfulness prediction and multimodal contrastive learning.

\paragraph{Review Helpfulness Prediction}
Customer reviews play an important role in helping customers investigate products before determining whether to purchase~\cite{zhu2010impact,diaz2018modeling,gamzu-etal-2021-identifying}.  
Support vector machine (SVM) was first employed to automatically judge the review helpfulness~\cite{kim2006automatically,zhang2006utility,tsur2009revrank}. Later, linear regression~\cite{lu2010exploiting,ghose2010estimating}, extended tensor factorization~\cite{moghaddam2012etf}, and probabilistic matrix factorization models~\cite{tang2013context} have been applied to integrate complicated constraints into the learning process. With the development of deep learning, deep neural networks~\cite{lee2014predicting,fan2018multi,chen2018cross} have been utilized to model the sophisticated elements in this task. 
Recently, \citet{qu2020category} proposed a graph neural network to capture the intrinsic relationship between the products and their reviews. 
However, most existing studies only focus on the text of reviews, neglecting the images that usually exist in online reviews. This paper takes advantage of the images and proposes a novel contrastive learning framework with a selective attention mechanism to learn expressive multimodal features.

\paragraph{Multimodal Representation Learning}
The foremost problem of multimodal tasks lies in multimodal representation learning~\cite{baltruvsaitis2018multimodal}. 
The concept of multimodal representation learning covers many techniques, such as multimodal fusion~\citep{vielzeuf2018centralnet,wang2020deep,mai2020modality,han2021bi}, multimodal contrastive learning~\citep{yuan2021multimodal,han2021improving}, etc.
Attention-based architectures are the basic routine in multimodal fusion, but the formulations are similar. 
In this paper, knowing about the particularity of MRHP and its dataset, we devise a novel attention mechanism to better aggregate information in textual data.
Additionally, we also upgrade the application of contrastive learning. Unlike the ordinary treatment that divides samples into positive and negative groups according to ``from myself'' or ``not from myself''~\citep{cui2020unsupervised,liang2020learning}, we extract contrastive pairs according to the natural correlation in the dataset and construct the framework of two feature spaces termed as domains.


%% file: Contents/Method.tex
\section{Method}
In this section, we first introduce the problem definition of \taskfullname~(\taskabbrname). 
Then we elaborate on the model architecture and processing pipeline of our method.

\begin{figure*}[ht]
    \centering
    \includegraphics[trim=0.5cm 0.5cm 1.5cm 2cm, width=0.92\textwidth]{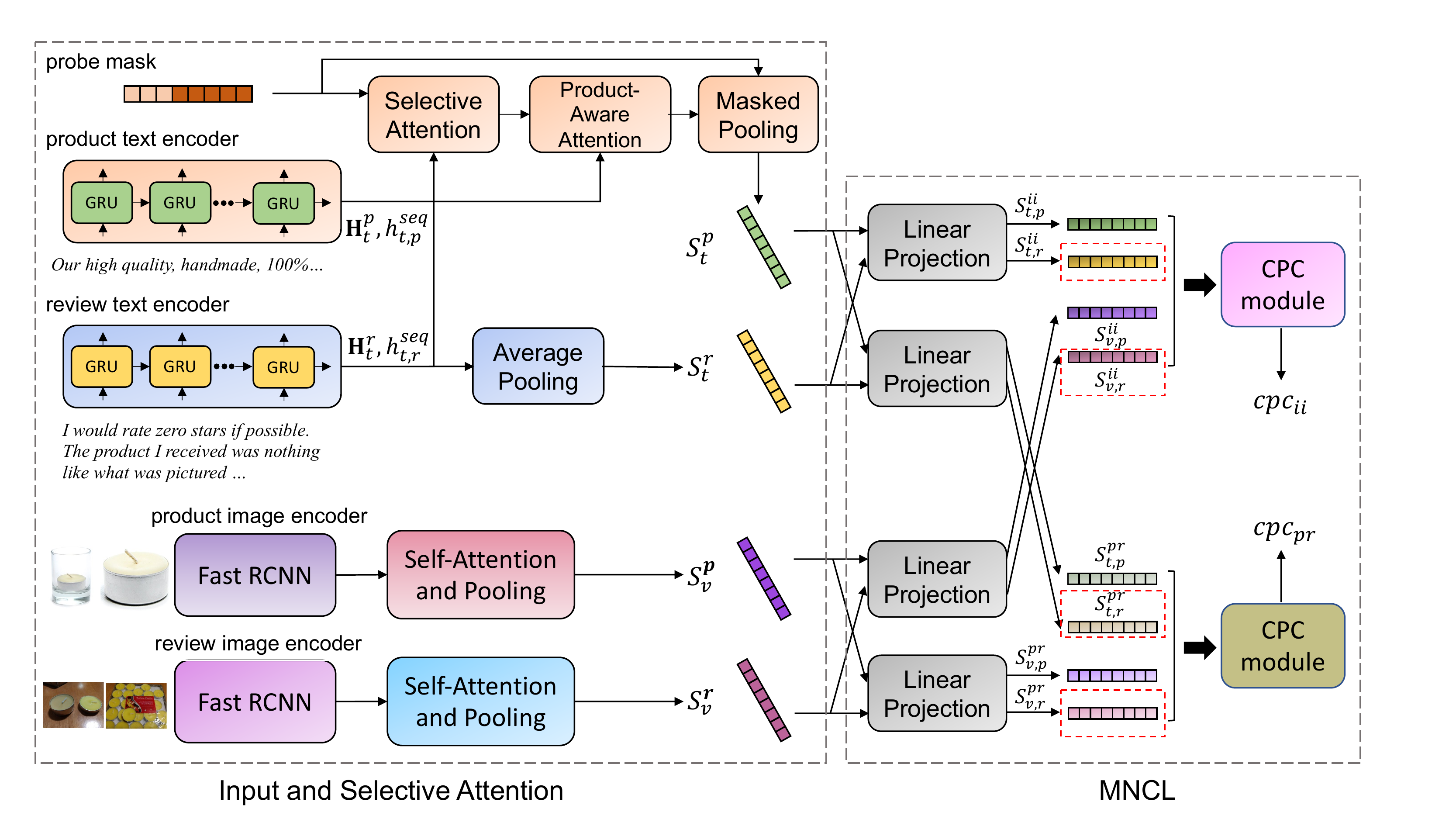}
    \caption{The overview of SANCL. The output layer is omitted. Features in red boxes ($S_{v,r}^{ii},S_{t,r}^{ii},S_{v,r}^{pr},S_{t,r}^{pr}$) are used in final helpfulness score prediction. }
    \label{model}
\end{figure*}

\subsection{Problem Definition}
Given a collection of product descriptions $\mathcal{P}=\{P_1,P_2,...,P_N\}$ and associated reviews $\mathcal{R}=\{R_1,R_2,...,R_N\}$ gleaned from an e-shopping website. Each product description $P_i\in \mathcal{P}$ contains the product name $n_{p_i}$ plus the text and image descriptions $T_{p_i}$ and $I_{p_i}$.
The underlying review collection $R_i$ associated with product $i$ contains $m$ review pieces $R_i=\{r_{i1},r_{i2},...,r_{im}\}$. 
Each review data frame is composed of images $I_{r_{ij}}$ and text $T_{r_{ij}}$ as well.
We exhibit an example of input data at the model's input position in Figure \ref{model}.
All review pieces are annotated with helpfulness scores $s_{ij}\in\{0,1,2,3,4\}$.
Multimodal review helpfulness prediction can be formulated as a regression task that aims to predict the helpfulness score of each review piece, and a ranking task to sort these reviews by their scores in descending order.
For the convenience of description, we call product description and review contents as \textit{fields}, while the data of text and image are termed as \textit{modalities} in the following sections.

\subsection{Overview}
The overall architecture of \modelname~is depicted in Figure \ref{model}.
We first generate a probe mask for each review according to the corresponding product name and review text as shown in Figure \ref{mask generation}. 
The probe mask highlights the sentences that mention the product, which then participates in the computation of selective attention to produce text representations.
For images, we feed the features extracted by pre-trained visual neural networks to two self-attention modules to produce image representations. 
Then we project these representations of each modality in both product description and customer review into two shared spaces (domains). 
We finally develop a contrastive learning module to compute the cross-modality and review-product contrastive scores, which further improves the quality of representations output from attention modules.

\subsection{Input Encoding}
\paragraph{Context-aware Textual Representation}
For both review and product text, we initialize their token representations with GloVe~\cite{pennington2014glove}\footnote{We used glove.840B.300d in our experiments (\url{https://nlp.stanford.edu/data/glove.840B.300d.zip}). } or pre-trained models as $\mathbf{E}_t = \{e_1^t, e_2^t, ..., e_l^t\} \in \mathbb{R}^{l\times d_e^t}$, where $l$ is the length (number of tokens) of a given sentence and $d_e^t$ is the embedding dimension. 
We then send these embeddings to a uni-directional Gated Recurrent Unit (GRU) \citep{cho2014learning}, yielding token-wise and sequence representations $\mathbf{H}_t= \{h_1^t,..., h_l^t\}$ and $h_t^{seq}$:
\begin{equation}
 \mathbf{H}_t, h_t^{seq} = \textbf{GRU}(\mathbf{E}; \theta_t).
\end{equation}
where $\theta_t$ is the parameters in GRU.

\paragraph{Visual Feature Extraction} We apply Faster R-CNN \citep{ren2015faster} on raw images and yield the hidden representations  $\mathbf{E}_v=\{e_1^v,e_2^v,...,e_n^v\} \in \mathbb{R}^{n\times d_e^v}$ in the last layer ahead of the classifier to map the Regions of Interest (RoI) in an image to a hidden space, where $n$ is the number of hot regions detected in the image and $d_e^v$ is the vector lengths of hidden representations. 
Same as \citet{liu2021multi}, we feed them into a self-attention module that outputs the encoded image representations $\mathbf{H}_v= \{h_1^v, h_2^v, ..., h_n^v\}$.

\subsection{Probe-based Selective Attention (PSA)}
\begin{figure}[ht]
    \centering
    \includegraphics[trim=1.5cm 5cm 11cm 0.5cm, width=\columnwidth]{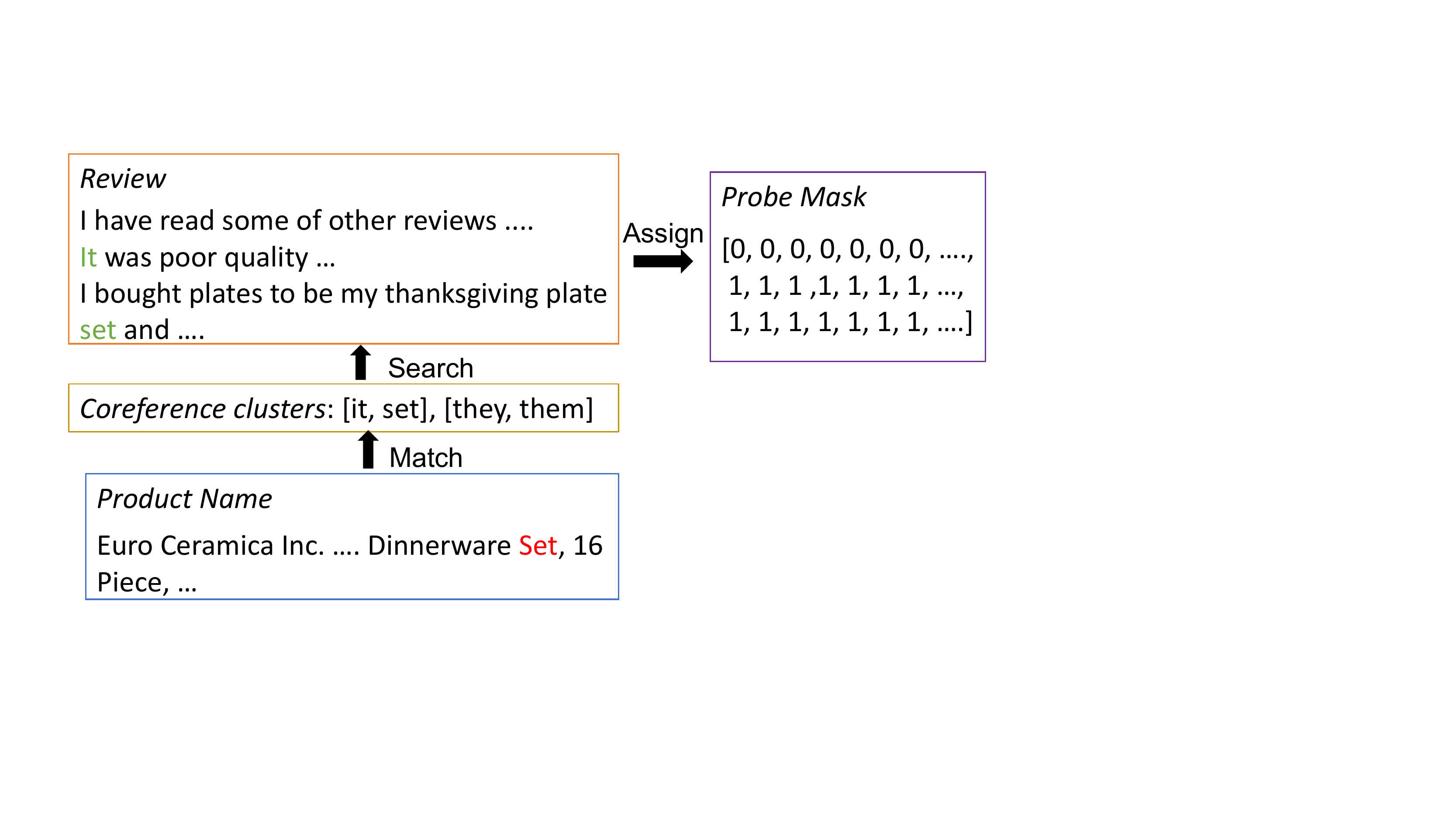}
    \caption{An example of mask generation}
    \label{mask generation}
\end{figure}

Previous work primarily formulated text attention as token-wise description--review attention
which fails to differentiate the relative importance among sentences and missing really task-related information.
Previous work fully relies on back-propagation to enforce the model to focus on those important token spans, but it is far from sufficient in this task---we desire to highlight valuable sentences which share the task-related information.
Intuitively, task-related information is more likely to exist in the sentences where a product or its properties, prices, etc., are mentioned. 
To this end, we generate a customized 'probe mask' for the review text to highlight sentence-level relevance.
\paragraph{Probe Mask Generation} The probe mask should reflect the position (i.e., in which sentence) where the product is mentioned in a review.
An example of the generation process is displayed in Figure \ref{mask generation}.
We first retrieve the core words from the product name by looking up its dependency tree and picking the lemmatized form of the words around the root. 
Next, we leverage an open-sourced coreference resolution package to identify all coreference clusters in the review. 

There are three possible resolution results: (1) A cluster containing the core word of the product name; (2) At least one cluster exists but the core word is missing in all clusters; (3) No coreference cluster exists. 
For (1) and (3), we do not require extra steps as the existence of entity clusters can be confirmed.
For (2), we are still uncertain whether an entity cluster is in the text. 
We devise a simple rule to tackle this situation---we regard the first cluster as the product name mention cluster, based on our observation that the first repeatedly mentioned pronouns in a review are more likely to refer to the product. After locating these product name mentions, we create the probe mask $M \in \mathbb{R}^{1\times l}$ by assigning 1 to the positions of those mention-found sentences and 0 to others. The process is summarized in Algorithm~\ref{alg:1}.

\input{Contents/submodules/algorithm1}

\paragraph{Selective Attention with Probe Mask} 
There are three steps to acquire product-aware review representations---self-attention, cross-field text attention (attention between product and review text), and pooling, among which the first and last steps take advantage of probe masks generated in the last step. 
We transform the binary probe masks into the real-value format:
\begin{equation}
    M' = \alpha M + \beta (1 - M),
\end{equation}
where $1 > \alpha > \beta > 0$, since we expect the mask could focus more on the sentences where the product is mentioned. 
This effect embodies in the self-attention computation of the review text $\mathbf{H}_t^r$, where the fundamental attention weights are computed as:
\begin{equation}
    \mathbf{A} = \mathbf{softmax}(\mathbf{WH}_t^r),
\end{equation}
We renew the original attention matrix $A \in \mathbb{R}^{l \times l}$:
\begin{equation}
    \mathbf{A}' = (M')^TM' \odot \mathbf{A},
\end{equation}
In this process, the attention weights are actually re-weighted as:
\begin{equation}
    a'_{ij} = \left\{
    \begin{aligned}
    \alpha^2a_{ij}, \; & \textrm{if}\; m_i = m_j = 1 \\
    \alpha\beta a_{ij}, \; & \textrm{if}\; m_i = 1, m_j = 0 \\
    \beta^2 a_{ij}, \; & \textrm{if}\; m_i=m_j=0
    \end{aligned}
    \right.,
\end{equation}
An intuitive explanation for this would be that a token receives more information from hot regions (whose mask value is $1$) than non-hot regions, and the relative impact power of these two regions is $\alpha/\beta$. 
Naturally, we set $\alpha$ to 1.0 while generating $\beta$ individually for each review from its sequence representation:
\begin{equation}
    \beta = \mathbf{sigmoid}(\mathbf{W}_{gen}h_{t,r}^{seq}+b_{gen}),
\end{equation}
where $\mathbf{W}\in \mathbb{R}^{d_h\times 1}$ is the weight matrix and $b \in \mathbb{R}$ is the bias. 
The sigmoid function ensures that $\beta \in (0,1)$.
Then we acquire the self-attention results as in common practice:
\begin{equation}
    \mathbf{H}_t^{r'}= \mathbf{H}_t^r + \mathbf{A}'\mathbf{H}_t^r,
\end{equation}
In cross-field text attention (i.e., the attention between product and review text), since weighted-sum is performed on the product text, we do not utilize probe masks in this stage and obtain $\mathbf{H}_t^{r''}$. 
Finally, we average the result with the probe mask by a weighted sum to aggregate these sentence representations:
\begin{equation}
    S_t^r = \mathbf{weighted\_sum}(\mathbf{H}_t^{r''}, M').
\end{equation}
Note that for image representations there are only cross-image attention and average pooling to yield $S_v^r$.

\subsection{Multi-domain Natural Contrastive Learning (MNCL)}
From the theory of mutual information, training to split positive samples from negative ones by their similarity can enrich the learned representations and enhance downstream tasks' performance. 
In our work, we are concerned about natural relations and split them into two domains: the inner-instance domain ($ii$) and the product (introduction)-review ($pr$) domain. 
Before forwarding the input representations into the MNCL module, all pooled representations are projected to the shared representation spaces of each domain through a projection network, which is composed of two linear layers with an activation layer in between.
We denote them as $S_{m,f}^d$, where $m \in \{t,v\}$ is the modality type, $f \in \{r,p\}$ is the field (review or product description) and $d \in \{ii, pr\}$ is the domain name:
\begin{equation}
    S_{m,f}^{d} = W_{m,f,2}^d\mathbf{Tanh}(W_{m,f,1}^dS_m^d+b^d_{m,f,1}) + b_{m,f,2}^d
\end{equation}
where $W_{m,*,i}^d$ and $b_{m,*,i}^d$ are weights and biases in the $i$-th layer of the projection network. Note that the data in the same modality and domain share the same network parameters.
In the succeeding content, we are going to describe details of the two contrastive-learning domains, mainly concerning how to pick positive and negative samples for contrastive learning and training.

\paragraph{Inner Instance (II) Domain}
In the inner instance domain, we separate positive and negative pairs according to how similar the representations between image and text are in a single training instance. 
First, from the sellers' perspective, the text and image of a product should match well so as to attract customers. 
Thus we mark text-image pairs of product descriptions as positive ones (the set of these pairs is denoted as $\mathbb{S}^p_{ii}$).
Besides, from our observation, reviews that achieve high helpfulness scores possess a high similarity between their text and the attached image. 
Therefore, we mark the former as positive (the set is denoted as $\mathbb{S}^+_{ii}$) and the latter as negative (the collection is denoted as $\mathbb{S}^-_{ii}$).

\paragraph{Product-Review (PR) Domain} 
The semantic matching property also exists between product descriptions and their associated reviews. 
As helpfulness is dependent on how well a review is pertinent to the theme of the product, we argue that review pieces of high helpfulness scores ($\mathbb{S}^+_{pr}$) should match the product introduction both visually and literally, while those low-score pieces ($\mathbb{S}^-_{pr}$) match the introduction poorly in both modalities.

\paragraph{Multi-domain Contrastive Predictive Coding (MCPC)}
In contrastive predictive coding \citep{oord2018representation}, we need to compute contrastive scores for every sample pair. 
According to the common approach \citep{yuan2021multimodal, han2021improving}, an exponential function is chosen as the score function:
\begin{equation}
    \varphi(\textbf{A},\textbf{B}) = \mathbf{exp}\left(\frac{\mathbf{norm}(\mathbf{A}^T)\mathbf{norm(B)}}{\tau}\right),
\end{equation}
where $\mathbf{norm(*)}$ is the l2-norm function, $\tau$ is the temperature hyper-parameter, for simplicity we keep its value 1.0 in our experiments. 
By noise contrastive estimation \citep{gutmann2010noise}, in the inner instance domain, the score is computed as:
\begin{multline}
    cpc_{ii} = -\sum_{(S_{t,j},S_{v,j})\in (\mathbb{S}_{ii}^+\cup\mathbb{S}^P_{ii})} \\
    \log\frac{\varphi(S_{t,j}, S_{v,j})}{\sum_{S_k\in (\mathbb{S}^+_{ii}\cup \mathbb{S}^-_{ii}\cup\mathbb{S}^P_{ii})}\varphi(S_{t,k}, S_{v,k})},
\end{multline}
where $(S_{t,j}, S_{v,j})$ are the text-image pair from the instance, i.e., a review piece or product description. The summation is over $\mathbb{S}^+_{ii}$ and $\mathbb{S}^P_{ii}$ because instances counted here are from both product descriptions and review pieces. Similarly in the product--review domain the score is:
\begin{multline}
    cpc_{pr}^m = -\sum_{S^r_{m,j}\in \mathbb{S}^+_{pr}} \\
    \log\frac{\varphi(S_{m,j}^r, S_{m,j}^p)}{\sum_{S_k\in (\mathbb{S}^+_{pr}\cup \mathbb{S}^-_{pr})}\varphi(S_{m,k}^r, S_{m,k}^p)}.
\end{multline}
\begin{equation}
    cpc_{pr} =  cpc_{pr}^t + cpc_{pr}^v
\end{equation}
where $S_{m,j}^r$ is the representation of modality $m$ in review $r$ from the positive review set $\mathbb{S}^+_{pr}$ and $S_{m,j}^p$ is the counterpart of the corresponding product. 

\subsection{Prediction and Training}
We select all review-related representations from the common spaces of two domains ($S_{t,r}^{ii}, S_{t,r}^{pr}, S_{v,r}^{ii}, S_{v,r}^{pr}$) and concatenate them as feature vectors for prediction ($\mathbf{F}$). 
A linear layer takes these feature vectors as input and outputs the helpfulness score predictions $\xi_r$:
\begin{equation}
    \mathbf{F} = \mathbf{concat}([S_{t,r}^{ii}, S_{t,r}^{pr}, S_{v,r}^{ii}, S_{v,r}^{pr}])
\end{equation}
\begin{equation}
    \xi_r = \mathbf{W}_o\mathbf{F}+b_o,
\end{equation}
where $\mathbf{W_o}$ and $b_o$ are the weight matrix and bias in the output layer.
Same as \citet{liu2021multi}, we adopt the standard pairwise ranking loss as the main task loss:
\begin{equation}
    \mathcal{L}_{task} = \sum_i \mathbf{max}(0, \gamma - \xi_{r^+,i} + \xi_{r^-,i}),
\end{equation}
where $r^+, r^-$ are an arbitrary positive/negative pair of review pieces under product $P_i$, $\gamma$ is a scaling factor. Contrastive losses make up the auxiliary loss:
\begin{equation}
    \mathcal{L}_{aux} = cpc_{ii} + cpc_{pr}
\end{equation}
The total loss for training is
\begin{equation}
    \mathcal{L} = \mathcal{L}_{task} + \kappa \mathcal{L}_{aux}
\end{equation}
where $\kappa$ is a hyper-parameter to adjust the effect of auxiliary loss.

%% file: Contents/submodules/algorithm1.tex
\begin{algorithm}[ht!]
\SetAlCapFnt{\small}
\SetKwFunction{Lemma}{Lemmatise}\SetKwFunction{Ext}{FindWordsNearRoot}
\SetKwFunction{Parse}{DependencyParse}\SetKwFunction{Init}{ZeroInit}
\SetKwFunction{Coref}{FindCoreferenceCluster} \SetKwFunction{Size}{size} \SetKwFunction{Return}{return}
\SetKwData{Gc}{gold\_cluster}
\SetKwArray{Clstr}{clusters}
\SetKwFunction{Start}{start}\SetKwFunction{End}{end}

\SetKwArray{mask}{mask} \SetKwData{sent}{sent}

\small
\SetAlgoLined
\KwIn{Review sentences $R$, product name $P$}
\KwOut{Probe mask $M$}
\BlankLine
\emph{\# core words and coreference clusters extraction}: \\
$\hat{R} \leftarrow \Lemma{R}, \hat{P}\leftarrow \Lemma{P}$\;
\emph{\# core words extraction}: \\
$T \leftarrow \Parse (\hat{P})$\;
$W \leftarrow \Ext{T}$\;
$\Clstr \leftarrow \Coref (\hat{R})$ \\

\BlankLine
\emph{\# mask generation}: \\
$M \leftarrow \Init{R.\Size}$ \\
\If{C = $\emptyset$}{
    {\bf return} $M$
}
\ForEach{c {\bf in} \Clstr}{
    \If{{\bf any} $w \in W$ {\bf in} c}{
        \Gc = c
    }
}
\If{\Gc $= \emptyset$}{\Gc = \Clstr{$0$}}
\ForEach {\sent $\in \hat{R}$}{
    \If{{\bf any} $w\in \Gc$ {\bf in} \sent}{
       $M[\sent.\Start : \sent.\End] \leftarrow True$ 
    }
}
{\bf return} $M$

\caption{\small Probe Mask Generation}
\label{alg:1}
\end{algorithm}

%% file: Contents/Experiments.tex
\section{Experimental Settings}
This section presents the datasets, baseline models, and metrics used and compared in our experiments. 

\input{Contents/submodules/results}

\subsection{Datasets}

We conduct experiments on three \taskabbrname~datasets \citep{liu2021multi} in different categories: \textit{Clothing, Shoes \& Jewelry, Home \& Kitchen} and \textit{Electronics}. 
The text and images in these datasets are crawled from a number of Amazon online shops under corresponding categories from the year 2017 to 2019.
The helpfulness scores equal to $\lfloor{\log_2 n_{\rm{votes}}}\rfloor$ and are then clipped into $[0,4]$. 
More details of datasets are provided in Appendix.

\subsection{Baseline Models}
Following previous work, we first compare our model with a bunch of baselines on the text-only setting, which examines the effect of our selective attention mechanism and text-related contrastive learning modules. 
The baseline candidates contain Multi-Perspective Matching (BiMPM) network~\citep{wang2017bilateral}, Embedding-gated CNN (EG-CNN)~\citep{chen2018cross}, Convolutional Kernel-based Neural Ranking Model (Conv-KNRM)~\citep{dai2018convolutional} and Product-aware Helpfulness Prediction Network (PRHNet)~\citep{fan2019product}.
In multimodal settings, we pick a collection of state-of-the-art multimodal helpfulness prediction models for comparison:
\begin{itemize}
    \item \textbf{SSE-Cross}~\citep{abavisani2020multimodal}: The Stochastic Shared Embeddings (SSE) Cross-modal Attention Network introduces a novel cross-attention mechanism that can filter noise components from weak modalities which may mislead the model to make wrong predictions on a sample. 
    SSE is adopted as the regularization technique to alleviate over-fitting in the fusion process to further prompt the prediction accuracy.
    \item \textbf{D\&R Net}~\citep{xu2020reasoning}: The Decomposition and Relation Network learns the commonality and discrepancy between image and text in the decomposition network and the multi-view semantic association information in the relation network.
    \item \textbf{MCR}~\citep{liu2021multi}: The Multi-perspective Coherent reasoning method incorporates the joint reasoning across textual and visual modalities from both the product and the review. Three types of coherence are modeled to learn effective modality representations for helpfulness prediction.
\end{itemize}
In both settings, we also test our method with BERT~\cite{devlin2018bert} as the text encoder. 
In addition, we test and record the BERT-with-head performance (BERT+a double linear layers) as the blank comparison experiment.

\subsection{Metrics}
We utilize several metrics for ranking tasks to evaluate the performance of these models.
After sorting all prediction scores by their corresponding truth scores in descending order, the Mean Average Precision (MAP) computes the mean precision till the sample with the $k$-th highest score. 
K is usually large enough to encompass the entire collection of reviews under every product.
The Normalized Discounted Cumulative Gain (NDCG-N) \citep{jarvelin2017ir,diaz2018modeling} purely reckons the gain value over top-N predictions (N is 3 and 5 in our experiments), which simulates the real circumstances of a typical customer who would always read the topmost reviews.

%% file: Contents/submodules/results.tex
\renewcommand{\arraystretch}{1.05}
\begin{table*}[ht]
    \small
    \resizebox{\textwidth}{!}{
    \begin{tabular}{l|l*{3}{|c c c}}
        \toprule
        \multirow{2}{*}{Setting} & \multirow{2}{*}{Model} & 
        \multicolumn{3}{c|}{Cloth. \& Jew} & \multicolumn{3}{c|}{Electronics}  & \multicolumn{3}{c}{Home \& Kitchen}   \\
        \cline{3-11}
        ~ & ~ & \textbf{MAP} & \textbf{N-3} & \textbf{N-5} & \textbf{MAP} & \textbf{N-3} & \textbf{N-5} & \textbf{MAP} & \textbf{N-3} & \textbf{N-5} \\
        \midrule
         \multirow{8}{*}{Text-only} & 
         BiMPM$^\ast$~\cite{wang2017bilateral} & 57.7 & 41.8 & 46.0 & 52.3 & 40.5 & 44.1 & 56.6 & 43.6 & 47.6 \\
        & EG-CNN$^\ast$~\cite{chen2018cross} & 56.4 & 40.6 & 44.7 & 51.5 & 39.4 & 42.1 & 55.3 & 42.4 & 46.7 \\  
        & Conv-KNRM$^\ast$~\cite{dai2018convolutional} & 57.2 & 41.2 & 45.6 & 52.6 & 40.5 & 44.2 & 57.4 & 44.5 & 48.4 \\
        & PRHNet$^\dagger$~\cite{fan2019product} &  58.23 & 43.36 & 47.21 & 52.31 & 40.43 & 43.88 & 57.11 & 44.46 & 48.27 \\
        & SANCL (Ours) & \textbf{58.98}$^{\natural}$ & \textbf{44.75}$^{\natural}$ & \textbf{48.57}$^{\natural}$ & \textbf{53.03}$^{\natural}$ & \textbf{41.03}$^{\natural}$ & \textbf{44.77}$^{\natural}$ & \textbf{58.03}$^{\natural}$ & \textbf{45.59}$^{\natural}$ & \textbf{49.31}$^{\natural}$ \\
        \cline{2-11}
        & BERT~\cite{devlin2018bert} & 56.47 & 42.98 & 46.84 & 51.95 & 39.77 & 43.11 & 56.62 & 42.12 & 46.87 \\
        & PRHNet+BERT$^\dagger$~\cite{fan2019product} &  57.51 & 43.65 & 47.74 & 52.28 & 40.66 & 44.02 & 57.32 & 44.74 & 48.42 \\
        & SANCL+BERT (Ours) & \textbf{58.49}$^{\natural}$ & \textbf{44.91}$^{\natural}$ & \textbf{48.69}$^{\natural}$ & \textbf{53.13}$^{\natural}$ & \textbf{41.77}$^{\natural}$ & \textbf{45.01}$^{\natural}$ & \textbf{58.20}$^{\natural}$ & \textbf{45.83}$^{\natural}$ & \textbf{49.65}$^{\natural}$\\
        \midrule
         \multirow{6}{*}{Multimodal} & SSE-Cross$^\ast$~\cite{abavisani2020multimodal} & 65.0 & 56.0 & 59.1 & 53.7 & 43.8 & 47.2 & 60.8 & 51.0 & 54.0 \\
        & D\&R Net$^\ast$~\cite{xu2020reasoning} & 65.2 & 56.1 & 59.2 & 53.9 & 44.2 & 47.5 & 61.2 & 51.8 & 54.6 \\
        & MCR$^\dagger$~\cite{liu2021multi} & 66.96 & 58.03 & 61.06 & 55.86 & 46.32 & 49.45 & 63.17 & 53.85 & 57.14  \\
        & SANCL (Ours) & \textbf{67.26} & \textbf{58.61}$^{\natural}$ & \textbf{61.48}$^{\natural}$ & \textbf{56.19} & \textbf{46.98}$^{\natural}$ & \textbf{49.92}$^{\natural}$ & \textbf{63.35} & \textbf{54.28}$^{\natural}$ & \textbf{57.40} \\
        \cline{2-11}
        & MCR+BERT~\cite{liu2021multi} & 65.81 & 55.94 & 58.75 & 55.15 & 45.67 & 48.62 & 62.39 & 52.91 & 56.09 \\
        & SANCL+BERT (Ours) & \textbf{66.52}$^{\natural}$ &\textbf{56.73}$^{\natural}$ & \textbf{59.90}$^{\natural}$ & \textbf{56.04}$^{\natural}$ & \textbf{46.77}$^{\natural}$ & \textbf{49.95}$^{\natural}$ & \textbf{62.74} & \textbf{53.65}$^{\natural}$ & \textbf{56.91}$^{\natural}$ \\
        \bottomrule
    \end{tabular}
    }
    \caption{Results on three datasets; all reported metrics are the average of five runs; ``$\ast$'' are from \citet{liu2021multi} and ``$\dagger$'' are from the open-source code in~\citet{liu2021multi}; ``$\natural$" represent the results significantly outperforms PRHNet and MCR with p-value < 0.05 based on paired t-test.}
    \label{amazon results}
\end{table*}

%% file: Contents/Analysis.tex
\section{Results and Analysis}
In this section, we will compare our approach with several advanced baselines and explore how it improves the multimodal helpfulness prediction task.

\subsection{Performance Comparison}
We list the performance of our model and baselines in Table \ref{amazon results}.
Notably, \modelname~consistently outperforms all the baselines in both text-only and multimodal settings and under both BERT and Glove initialization methods.
These outcomes initially demonstrate the efficacy of our method in \taskabbrname~tasks.
It is surprising that we cannot gain a significant performance boost by replacing Glove with BERT as the text encoder.
We speculate the reason is that Glove embeddings are expressive enough for this task.

Moreover, it can be claimed that \modelname~is a lightweight model compared to the multimodal SOTA since the model size and GPU memory consumption of \modelname~are much lower than MCR.
The total number of parameters is 2.63M in MCR and 1.41M (excluding the embedding layer) in \modelname~respectively, which indicates a double efficiency.
The average GPU memory usage of \modelname~during the training on Amazon-MRHP Home \& Kitchen is around 2.4G, while MCR occupies an average of 13.7G GPU memory during training, which is 4.7 times higher than \modelname. 

\input{Contents/submodules/ablation}
\subsection{Ablation Study}
To verify the benefits of our proposed method, we carry out comprehensive ablation experiments on the Amazon electronics dataset, including the selective attention and contrastive learning components.
In selective attention, we first replace learned $\beta$ with a fixed value of 0.5, since we find most $\beta$ values in our experiments are around 0.5. Next, we remove the entire selective attention module and only preserve the primitive attention computation.
The decline in the outcome of both situations manifests that probe-based selective attention amends the cross-text information exchange between text fields.
For multi-domain contrastive learning, we delete the CPC losses of a single or both domains in training. 
The results indicate that both domains have a positive impact on performance. 
Moreover, the effect of the two domains does not counteract their collaboration, as we observe accumulated benefits when they operate together.

\subsection{Case Study}
To understand how our model deals with samples in-depth, we randomly pick up a test product-review instance from the test set of Amazon Home \& Kitchen to explain how SANCL works, as shown in Table \ref{Case Study}. 

\begin{figure}[ht]
    \centering
    \includegraphics[trim=2cm 0cm 2cm 0cm, width=0.45\columnwidth]{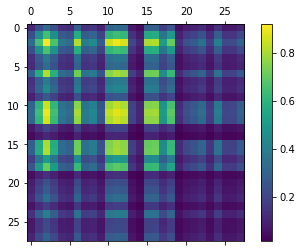}
    \caption{The self-attention weights of the review text in the given example at test time ($\beta$=0.57).}
    \label{attention weights}
\end{figure}

In this example, the customer bought the pins to fix the edge of his sofa. 
Instead of photoing pins themselves, the customer only presented the tidy sofa after installing the pins.
We first visualize the attention weights in test time, as shown in Fig. \ref{attention weights}.
Note that only the first sentence in the review contains the elements in the coreference clusters, which we have emphasized with italics and underlined in Table \ref{Case Study}.
Consistently, we observe the significantly larger weights in the region of the first sentence (row/column 1-19) while the rest region's weights are much smaller.
We also ran MCR and collect its prediction on this example, and it is clear that MCR commits a severe error here, probably caused by the direct classification of the unimodal cosine similarity.
In our approach, as we carefully analyze and classify the positive and negative pairs in the multi-domain contrastive learning framework, the huge semantic similarity between review text and image and between product description and review text, indicated by the high CPC scores $S_{v,r}^{ii}$, assists the model to correctly predict the score.

\input{Contents/submodules/case_study}

%% file: Contents/submodules/ablation.tex
\begin{table}[th]
  \small
    \centering
    \resizebox{\linewidth}{!}{
    \begin{tabular}{l|*{3}{c}}
        \toprule
        Description & \textbf{MAP} & \textbf{N-3} & \textbf{N-5} \\
        \midrule
        \modelname\ & \textbf{56.19} & \textbf{46.98} & \textbf{49.92} \\
        \midrule
        Attention \\
        \quad w/o learned $\beta$ (fixed at 0.5) & 55.61 & 46.37 & 49.58 \\
        \quad w/o probe mask & 55.43 & 46.11 & 49.45 \\
        \midrule
        Contrastive learning \\
        \quad w/o $cpc_{ii}$ & 55.54 & 46.29 & 49.23 \\
        \quad w/o $cpc_{pr}$ & 55.81 & 46.40 & 49.47 \\
        \quad w/o $cpc_{ii}$ and $cpc_{pr}$ &  55.35 & 46.28 & 49.09 \\
        \bottomrule
    \end{tabular}
    }
    \caption{Ablation study of \modelname~on the Electronics dataset.}
    \label{Abl Study}
\end{table}

%% file: Contents/submodules/case_study.tex
\begin{table}[ht]
    \centering
    \small
    \resizebox{\linewidth}{!}{
    \begin{tabular}{p{\linewidth}}
        \toprule
        \textbf{Product Name}: Twisty Pins for Upholstery, Slipcovers, and Bedskirts 50/pkg\\
        \midrule
        \textbf{Product description}: Package of 50 Clear Twisty Pins for securing fabrics and accent trims. Nickel-plated steel pin 1/2" in diameter clear top, wire twist 3/8" long. Perfect for Medium to lightweight fabrics, bed skirts, bed ruffles, slipcovers, and upholstery.
        
        \subfigure{\includegraphics[height=\csfigheight]{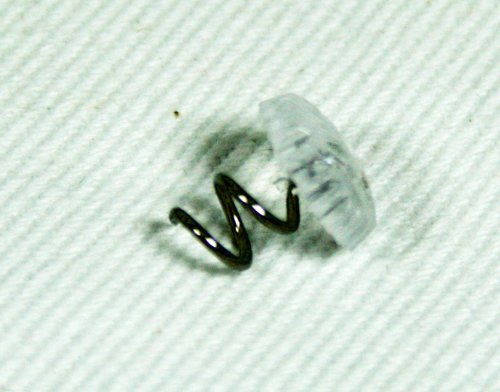}}
        \subfigure{\includegraphics[height=\csfigheight]{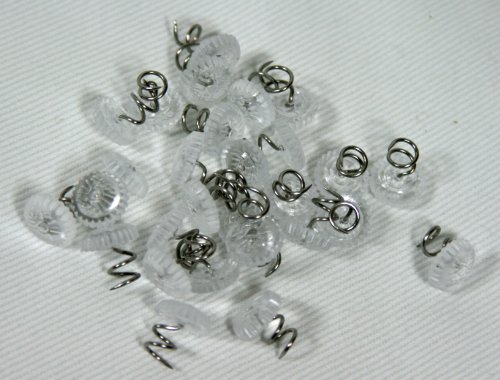}}
        \subfigure{\includegraphics[height=\csfigheight]{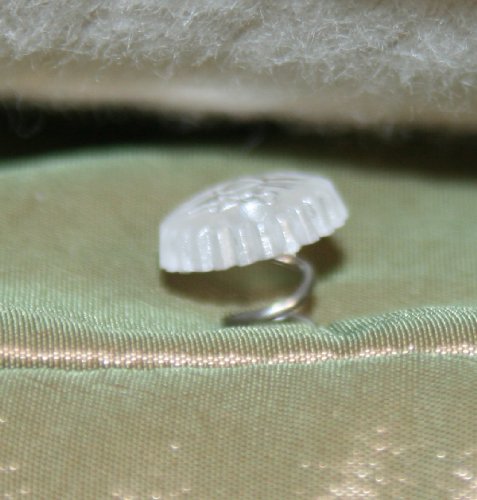}}
        \subfigure{\includegraphics[height=\csfigheight]{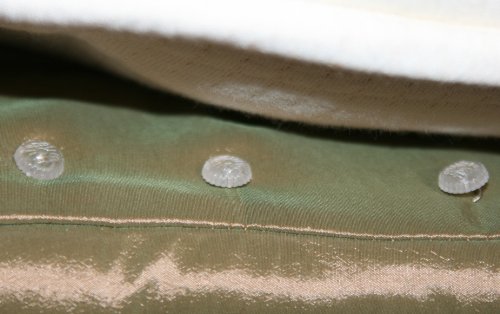}} \\
        \midrule
        
        \textbf{Review (Helpfulness Score: 4)}: I bought \underline{\textit{these}} to pin the loose material on a sofa cover and they worked like a charm. The sofa cover definitely looks form fitting now.
        
        \subfigure{\includegraphics[height=\csfigheight]{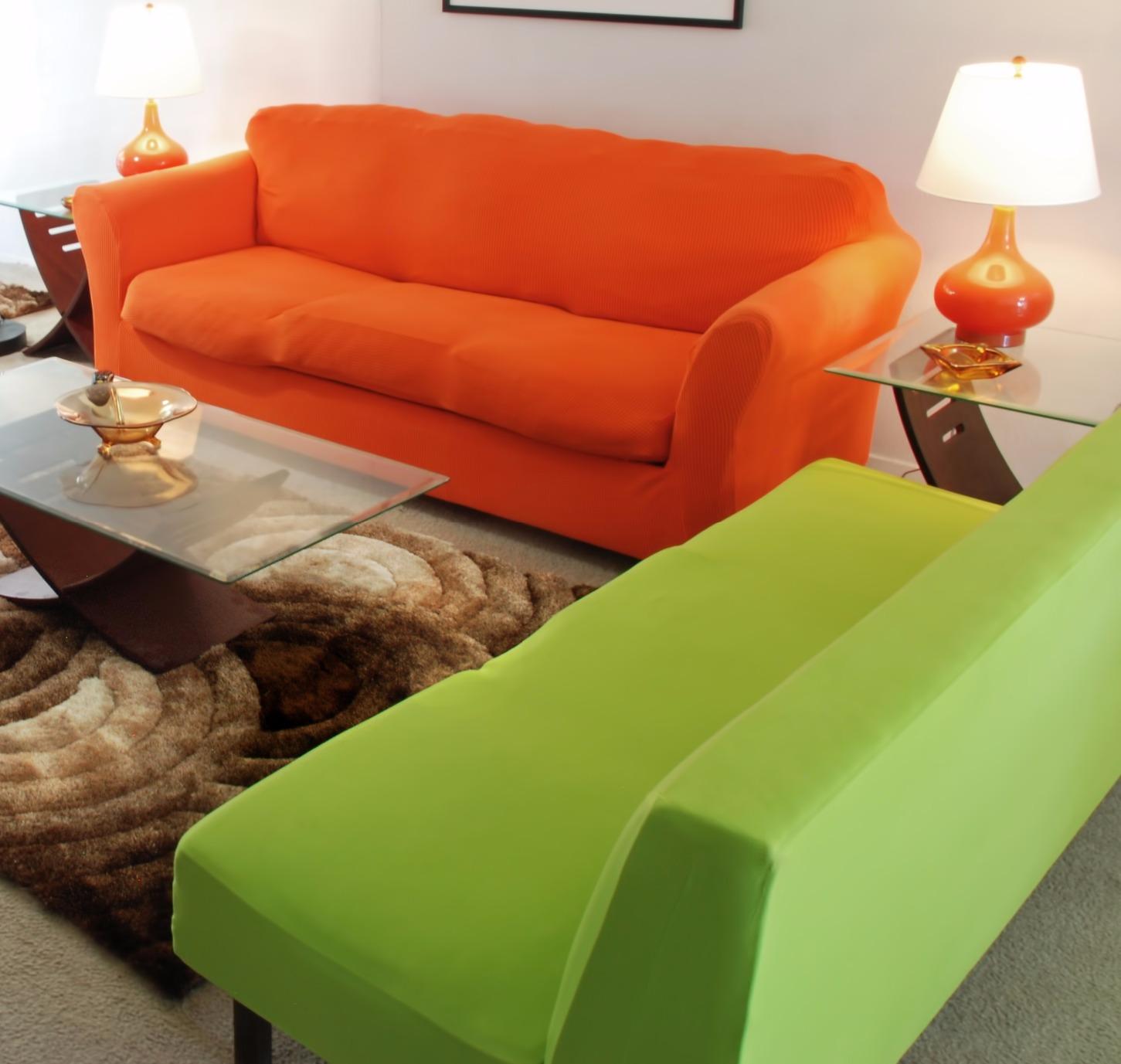}} \\
        \midrule
        \textbf{Predictions}: \quad \modelname: 4.5291 \quad MCR: -1.0832 \\
        \midrule
        \textbf{CPC score}: $cpc_{ii}=0.82, cpc_{pr}^t=0.76, cpc_{pr}^v=0.21$\\
        \bottomrule
    \end{tabular}}
    \caption{Examples from the Amazon Home \& Kitchen test set.}
    \label{Case Study}
\end{table}

%% file: Contents/Conclusion.tex
\section{Conclusion}
We propose a novel framework, \modelname, for the task of multimodal review helpfulness prediction (MRHP) in this paper.
We first present a selective attention mechanism, which purposefully aggregates information from these crucial sentences in the review text by generating the probe mask that exerts re-normalization on the attention weights and pooling stage.
We then build up a multi-domain natural contrastive learning framework in our model. 
It exploits the natural relations among the data from different fields and modalities in the dataset to enhance the model's capacity for multimodal representation learning.
Results of comprehensive experiments and analyses demonstrate the superiority of our model over the comparable baselines and the efficacy of the novel components.

\section*{Acknowledgement}
This work is supported by the A*STAR under its RIE 2020 AME programmatic grant
RGAST2003 and project T2MOE2008 awarded by Singapore's MoE under its Tier-2 grant scheme.

%% file: Contents/Appendix.tex
\newpage
\appendix

\section{Dataset Specification}
Specifications of Amazon-MRHP are listed below.
\input{Contents/submodules/datasets}

\section{Hyperparameter Search}
The optimal hyperparameter settings are provided in Table \ref{Amazon hyperparameters} and \ref{Amazon Bert hyperparameters}.

\begin{table}[h!]
    \centering
    \resizebox{\linewidth}{!}{\begin{tabular}{lccc}
    \toprule
    \multicolumn{4}{c}{\textbf{Glove Hyperparameters}} \\
         & Cloth. \& Jew. &  Elec. & Home \& Kitch. \\
         \midrule
         learning rate  & $1e^{-4}$  & $5e^{-5}$ & $1e^{-4}$ \\
         text embedding dim & 300 & 300 & 300 \\
         text embedding dropout & 0.5 & 0.5 & 0.2 \\
         image embedding dim & 128 & 128 & 128 \\
         LSTM hidden dim & 128 & 128 & 128 \\
         shared space hidden & 64 & 64 & 64 \\
         $\kappa$ & 0.25 & 0.1 & 0.1 \\
         batch size & 32 & 32 & 32 \\
         \bottomrule
    \end{tabular}
    }
    \caption{Hyperparameters for all categories using glove-300d embeddings.}
    \label{Amazon hyperparameters}
\end{table}

\begin{table}[h!]
    \centering
    \resizebox{\linewidth}{!}{\begin{tabular}{lccc}
    \toprule
    \multicolumn{4}{c}{\textbf{Amazon-MRHP Hyperparameters}} \\
         & Cloth. \& Jew. &  Elec. & Home \& Kitch. \\
         \midrule
         learning rate  & $2e^{-5}$  & $2e^{-5}$ & $2e^{-5}$ \\
         text embedding dim & 768 & 768 & 768 \\
         text embedding dropout & 0.5 & 0.5 & 0.5 \\
         LSTM hidden dim & 128 & 128 & 128 \\
         image embedding dim & 128 & 128 & 128 \\
         shared space hidden & 64 & 64 & 64 \\
         $\kappa$ & 0.3 & 0.25 & 0.25 \\
         batch size & 32 & 32 & 32 \\
         \bottomrule
    \end{tabular}
    }
    \caption{Hyperparameters for all categories using BERT as encoder}
    \label{Amazon Bert hyperparameters}
\end{table}

We use the same set of settings for text-only and multimodal modes for the same category dataset.
The search space of these hyperparameters are: learning rate in $\{1e^{-4}, 2e^{-5}\}$, text embedding dropout in $\{0.2, 0.5\}$, $\kappa$ in $\{0.1, 0.25, 0.3, 0.5\}$, shared space hidden dimension in $\{64, 128\}$.
We train and test each dataset on a single Tesla V100 GPU.
In BERT experiments, we use shared a BERT encoder for both product description and review text.
To balance the computation cost and model performance, following \citet{sun2019fine}, we fine-tune the last four layers of the BERT encoder.
\label{sec:appendix}

\section{Language Tools}
For coreference resolution, we use neuralcoref\footnote{\url{https://github.com/huggingface/neuralcoref}}, an extension that can be placed on SpaCy processors. For BERT model, we use the huggingface\footnote{\url{https://huggingface.co/docs/transformers/model\_doc/bert}} transformers package to load.

%% file: Contents/submodules/datasets.tex
\begin{table}[ht]
    \centering
    \small
    \resizebox{\linewidth}{!}{
    \begin{tabularx}{0.5\textwidth}{lccc}
        \toprule
        \multicolumn{4}{c}{\textbf{Amazon-MRHP (Products/Reviews)}} \\
        [0.5ex]
        Category & Cloth. \& Jew. & Elec. & Home \& Kitch.\\
        \midrule
        Train & 12074/277308  & 10564/240505 & 14570/369518 \\
        Dev &  3019/122148 & 2641/84402 & 3616/92707 \\
        Test & 3966/87492 & 3327/79750  &  4529/111593 \\
        \bottomrule
    \end{tabularx}}
    \caption{Statistics of the Amazon-MRHP dataset.}
    \label{Amazon-Spec}
\end{table}